# Composition of kernel and acquisition functions for High Dimensional Bayesian Optimization

Antonio Candelieri[1][0000-0003-1431-576X], Ilaria Giordani[2][0000-0002-6065-0473], Riccardo Perego[2][0000-0003-0117-2237] and Francesco Archetti[2][0000-0003-1131-3830]

[1] Department of Economics, Management and Statistics, University of Milano-Bicocca, 20126, Milan, Italy

[2] Department of Computer Science, Systems and Communication, University of Milano-Bicocca, 20126, Milan, Italy

`antonio.candelieri@unimib.it`

**Abstract.** Bayesian Optimization has become the reference method for the global optimization of black box, expensive and possibly noisy functions. Bayesian Optimization learns a probabilistic model about the objective function, usually a Gaussian Process, and builds, depending on its mean and variance, an acquisition function whose optimizer yields the new evaluation point, leading to update the probabilistic surrogate model. Despite its sample efficiency, Bayesian Optimization does not scale well with the dimensions of the problem. The optimization of the acquisition function has received less attention because its computational cost is usually considered negligible compared to that of the evaluation of the objective function. Its efficient optimization is often inhibited, particularly in high dimensional problems, by multiple extrema. In this paper we leverage the additionality of the objective function into mapping both the kernel and the acquisition function of the Bayesian Optimization in lower dimensional subspaces. This approach makes more efficient the learning/updating of the probabilistic surrogate model and allows an efficient optimization of the acquisition function. Experimental results are presented for real-life application, that is the control of pumps in urban water distribution systems.

## 1 Introduction:

Bayesian Optimization (BO) [1,2] has become the reference method for the global optimization of a black box function, expensive and possibly noisy functions. BO maintains a probabilistic surrogate model – usually a Gaussian Process (GP) – of the objective function, updated depending on data as we observe them, namely function evaluations. GP-based BO has two key design factors: the choice of the GP kernel – which sets a prior about the hypothetical smoothness of the black-box function to optimize – and the acquisition function whose optimization provides the next points where to evaluate the objective function. Every new function evaluation is exploited to update the GP model.

BO has been generalized to the constrained case [3,4,5], including black-box constraints and more recently to partially defined objective functions or non-computable domains [6].

BO is sample efficient when applied to low dimensional problem, up to 10 or 20, but already midsize problems are challenging. This difficulty is of course exacerbated when the evaluations of the objective function are very expensive, like in most simulation-optimization problems, hyperparameter optimization of large-scale Machine Learning applications (e.g., Automated Machine Learning [7], Neural Architecture Search [8] and complex Machine Learning pipelines [9]). Many approaches have been suggested to mitigate the computational load of high dimensional BO, such as random embeddings [10] and the exploitation of the *additivity* of the objective function. [11,12] describe a method to search over possible kernel compositions starting with basic kernel. A major step was taken in [13] and successively developed in [14] by introducing, based on the Bochner's theorem, a Fourier Features approximation which is shown to yield an error decreasing exponentially with the number of features, while allowing a reduction in the complexity of the kernel matrix inversion.

In this paper we focus on a strategy which leverages additionality of the objective function into the decomposition of the kernel and the objective function. The main contributions of this paper are:

- To show that the function additivity can be leveraged to *(i)* reduce the cost of updating the GP and *(ii)* improve the performance of different acquisition functions.
- To show, in a realistic problem, that also the case of black-box constraints fits naturally into the proposed framework.
- To compare the proposed additive BO with the standard one providing results on a benchmark test function and a real-life case study.

## 2   Background

The global optimization problem is usually defined as:

$$\min_{x \in X \subset \mathbb{R}^d} f(x)$$

where the search space $X$ is generally box-bounded. A GP is a stochastic process whose all finite marginals are Gaussians, fully characterized by its domain $D \subseteq \mathbb{R}^d$, its prior mean (assumed to be zero here), and its kernel function $k: X \times X \to \mathbb{R}$:

$$f(x) \sim \mathcal{N}(\mu(x), \sigma^2(x))$$

where $\mu(x)$ and $\sigma^2(x)$ are the mean and the variance functions.

The covariance structure of the stochastic process is governed by the kernel function $k(x, x')$. Many kernels are available, such as Squared Exponential (SE) and Matérn among others [1,2]. The SE kernel has been used in this paper:

$$k(x, x') = e^{-\frac{\|x - x'\|^2}{2\ell^2}}$$

Let $D_{1:t} = \{(x_i, y_i)\}_{i=1,\ldots,t}$, where $y_t = f(x_t) + \varepsilon$, and $\varepsilon \sim \mathcal{N}(\mu_\varepsilon, \sigma_\varepsilon^2)$ in the case of a noisy objective function. It is then easy to derive an expression for the predictive distribution:

$$P(f_{t+1}|D_{1:t}, x_{t+1}) = \mathcal{N}(f_{t+1}|\mu_t(x_{t+1}), \sigma_t^2(x_{t+1}))$$

with

$$\mu_t(x_{t+1}) = \mathbf{k}^T [K + \sigma_\varepsilon^2 I]^{-1} \{f_1, \ldots, f_t\}$$

$$\sigma_t^2 = k(x_{t+1}, x_{t+1}) - \mathbf{k}^T [K + \sigma_\varepsilon^2 I]^{-1} \mathbf{k}$$

The acquisition function is the mechanism to implement the trade-off between exploration and exploitation in BO. Any acquisition function aims to guide the search of the optimum towards points with potentially low values of objective function either because the prediction of $f(x)$ is low or the uncertainty is high (or both). While *exploiting* means to consider the region of the search space providing more chance to improve the current best solution (with respect to the current surrogate model), *exploring* means to move towards less explored regions. Many acquisition functions have been proposed, such as Probability of Improvement, Expected Improvement, Confidence Bound (Upper/Lower Confidence Bound for maximization/minimization problems, respectively), Entropy Search, Predictive Entropy Search, Knowledge Gradient and Thompson Sampling (TS) – a brief review is provided in [2]. Each acquisition offers its own blend of the GP's mean and variance. One of the most widely adopted is the Upper/Lower Confidence Bound (UCB/LCB), respectively for solving maximization/minimization problems. The next promising point where to evaluate the objective function is obtained by solving the following auxiliary problem (+/− refer to UCB and LCB, respectively):

$$x_{t+1} = \underset{x \in D}{\operatorname{argmax}} \, \mu_t(x) \pm \beta_t^{1/2} \sigma_t(x)$$

where $\beta_t$ satisfies the converge criteria analysed in [15].

## 3    Additive functions and basic decomposition in BO

Additive GP based models assume that the objective function $f(x)$ is a sum of functions $f^{(j)}$ defined over low-dimensional components:

$$f(x) = \sum_{j=1}^{G} f^{(j)}(x^{(j)})$$

where each $x^{(j)}$ belongs to a low dimensional subspace $X^{(j)} \subseteq D$ and $G$ denotes the number of these components. We assume in this paper that $X^{(j)} \cap X^{(k)} = \emptyset$ if $k \neq j$.

The concept of additive functions translates to GPs, where the stochastic process is a sum of stochastic processes, each one having low dimensional indexing. The effective dimensionality of the model is defined as the largest dimension among all additive groups, $\bar{d} = max_{j \in [G]} \dim(X^{(j)})$.

Under the additive assumption, the kernel and the mean function of a GP decompose similarly to the GP's stochastic process. Specifically, $k(x, y) = \sum_{j=1}^{G} k^{(j)}(x^{(j)}, x'^{(j)})$ and $\mu(x) = \sum_{j=1}^{G} \mu^{(j)}(x^{(j)})$. Consequently, BO can be applied, independently, on each subspace, while at the same time including the cross correlation of additive groups through the observations $D_{1:t}$ restore a global look of the problem.

Following, we report the algorithm for additive BO with TS as acquisition function. In the case that a different acquisition function is used, such as EI or UCB/LCB, step 3 is not needed and step 5 is modified with the corresponding equation. Preliminary we provide some relevant notations. A kernel can be represented as $k(x, x') = \langle \Phi(x)^\intercal \Phi(x') \rangle$ where $\Phi(x)$ is a feature map function, with $\Phi(x) \in \mathbb{R}^m$ and $m \ll d$.

Thus, the covariance matrix can be written as $K_t = \Phi(X_t)^\intercal \Phi(X_t)$, with $X_t = \{x_1, \dots, x_t\}$. Let denote with $\nu_t = [\Phi(X_t)^\intercal \Phi(X_t), +\sigma_\varepsilon^2 I]^{-1} \{y_1, \dots, y_t\}$, then the approximated mean and variance of the GP can be computed as: $\tilde{\mu}_t(x) = \Phi(x)^\intercal \nu_t$ and $\tilde{\sigma}_t(x)^2 = \sigma_\varepsilon^2 \Phi(x)^\intercal [\Phi(X_t)^\intercal \Phi(X_t), +\sigma_\varepsilon^2 I]^{-1} \Phi(x)$, when $\|\Phi(x)\|_2 = 1$ (which is true for Random as well as Quadrature Fourier Features approximations [14]).

---
**Algorithm 1**

---
1: **for** $t = 1, \dots, T$ **do**
2:    Update $\nu_t$ and $\Phi(X_t)^\intercal \Phi(X_t) + \sigma_\varepsilon^2 I$
3:    Sample $\theta_t \sim \mathcal{N}(\nu_t, [\Phi(X_t)^\intercal \Phi(X_t) + \sigma_\varepsilon^2 I]^{-1})$
4:    **for** $j = 1, \dots, G$ **do**
5:       Find $x_{t+1} = \underset{x \in D}{\mathrm{argmax}}\, \theta_t^{(j)\intercal} \Phi^{(j)}(x^{(j)})$
6:    **endfor**
7:    Evaluate $f(x_{t+1}) = \sum_{j=1}^{G} f^{(j)}(x_{t+1}^{(j)})$
8: **endfor**

---

## 4 Experiments

### 4.1 Benchmark Function

The first experiment is related to a well-known test function, namely the Styblinski-tang [16]:

$$f(x) = \tfrac{1}{2} \sum_{i=1}^{d}(x_i^4 - 16 x_i^2 + 5 x_i)$$

where $d$ is the dimensions of the search space, which is the hypercube $x_i \in [-5; 5]$ for all $i = 1, \dots, d$. In our experiment we considered $d = 10$, where the maximum number of dimensions for each group of decision variables is $\bar{d} = 1$.

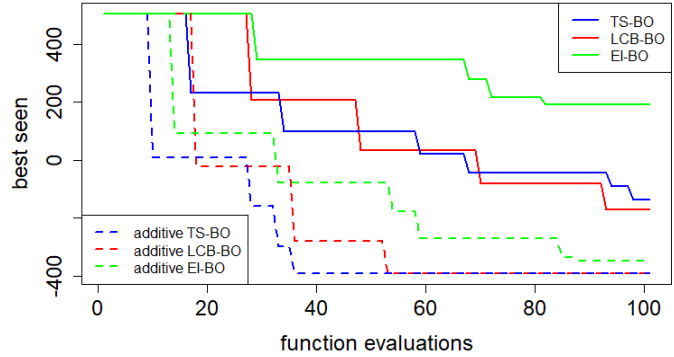

**Fig. 1.** Best-seen over function evaluations; comparison between additive BO and "standard" BO with three acquisition functions (i.e., TS, LCB and EI)

Figure 1 shows the best value of the objective function observed along the BO function evaluations (aka "best seen"). All the additive BO implementations resulted both more effective (i.e., they provided a better optimal solution) and more efficient (i.e., they achieve the optimal solution in a lower number of function evaluations) than "standard" BO. Moreover, TS and LCB resulted better than EI, irrespectively to standard and additive BO.

### 4.2 Control problems

Here we consider the Pump Scheduling Optimization (PSO) of an urban Water Distribution Network (WDN), using a benchmark WDN, namely AnyTown [17], powered by 4 variable speed pumps. The goal of PSO is to minimize the energy cost associated to the pump schedule, that is defined as the status (or speed) of the pumps at given times (e.g., hours of the day). Status refers to ON/OFF pumps, while speed refers to variable speed pumps.

$$\min C = \Delta t \sum_{t=1}^{T} c_E^t \sum_{k_v=1}^{N_v} \gamma\, x_k^t\ Q_k^t\ \frac{H_k^t}{\eta_k}$$

where $C$ is the total energy cost, $\Delta t$ is the time step, $C_E^t$ is the energy price at time $t$, $\gamma$ is the specific weight for water, $Q_{k_j}^t$ is the water flow provided by the pump $k_j$ at time $t$, $H_{k_j}^t$ is the head loss on pump $k_j$ at time $t$, $\eta_{k_j}$ is the efficiency of pump $k_j$, and $P_i^t$ is the pressure at the control node $i$ at time $t$, $N_v$ is the number of variable speed pumps.

The decision variables are $x_{k_v}^t \in (0,1)$, where each one represents the speed of pump $k_v$ at time $t$. Both energy cost and hydraulic feasibility of the a given pump schedule are computed via software simulation, more specifically through the tool EPANET.

In the case of non-additive BO, the problem has been solved in [17], where the search space of pump schedules had dimensionality $d = 96$ (i.e., 4 pump's speeds for each hour of the day). By using additivity, we solved 24 problems with dimensionality 4 each (i.e., $\bar{d} = 4$). Table 1 summarizes the preliminary results obtained on this benchmark study, compared to those obtained via "standard" BO (i.e., non-additive) and previously reported in [17]. Results confirm what observed on the test function experiment: additive BO provides better optimal solutions with less computational time.

**Table 1.** PSO of a WDN: results on the AnyTown benchmark, standard vs additive BO

| Strategy | Energy cost ($) | Iteration number | Overall clock time |
|---|---|---|---|
| BO-LCB [17] | 653.55 | 356 | 21256.48 [secs] |
| BO-EI [17] | 609.17 | 290 | 23594.36 [secs] |
| Additive BO-LCB | 605.23 | 344 | 12331.65 [secs] |
| Additive BO-EI | 653.55 | 276 | 10775.44 [secs] |
| Additive BO-TS | **604.51** | 281 | 11331.31 [secs] |

## 5 Conclusions

We have shown how BO performance can be boosted exploiting the additive property inherent in the objective function. This improvement is observed for several acquisition functions but is more significant for Thompson Sampling than LCB/UCB and Expected Improvement (EI). We observe experimentally that the additive algorithm convergence, proved for the cumulative regret, is no-regret also for the performance metric based on best seen.

Significantly this result translates from the test function to the real-life case of the control of a water distribution network. The resulting optimization problem has close to 96 variables: the additive algorithm brings it to a dimension (i.e., $\bar{d} = 4$) where BO is very sample efficient.

## 6 Acknowledgments

This study has been partially supported by the Italian project "PerFORM WATER 2030" – programme POR (Programma Operativo Regionale) FESR (Fondo Europeo di Sviluppo Regionale) 2014–2020, innovation call "Accordi per la Ricerca e l'Innovazione" ("Agreements for Research and Innovation") of Regione Lombardia, (DGR N. 5245/2016 - AZIONE I.1.B.1.3 – ASSE I POR FESR 2014–2020) – CUP E46D17000120009.
We greatly acknowledge the DEMS Data Science Lab for supporting this work by providing computational resources.